\documentclass[submission,copyright,creativecommons]{eptcs}
\providecommand{\event}{SCAV 2018} 
\usepackage{epsfig} 
\usepackage{amsmath} 
\usepackage{amssymb} 
\usepackage{enumerate}
\usepackage{subcaption}
\usepackage{xspace}
\usepackage{colortbl}
\usepackage[font=scriptsize]{caption}
\usepackage{wrapfig}
\usepackage{float}

\title{The CAT Vehicle Testbed: A Simulator with Hardware in the Loop for Autonomous Vehicle Applications}
\author{Rahul Kumar Bhadani
\institute{Department of Electrical and Computer Engineering\\
University of Arizona\\
Tucson, USA}
\email{rahulbhadani@email.arizona.edu}
\and
Jonathan Sprinkle \qquad\qquad Matthew Bunting
\institute{Department of Electrical and Computer Engineering\\University of Arizona\\
Tucson, USA}
\email{\quad sprinkjm@email.arizona.edu \quad\qquad mosfet@email.arizona.edu}
}
\def\titlerunning{CAT Vehicle Testbed: A Simulator with HIL for Autonomous Vehicle Applications}
\def\authorrunning{R.  Bhadani, J. Sprinkle \& M. Bunting}
\begin{document}
\maketitle

\begin{abstract}
This paper presents the CAT Vehicle (Cognitive and Autonomous Test Vehicle) Testbed: a research testbed comprised of a distributed simulation-based autonomous vehicle, with straightforward transition to hardware in the loop testing and execution, to support research in autonomous driving technology. The evolution of autonomous driving technology from active safety features and advanced driving assistance systems to full sensor-guided autonomous driving requires testing of every possible scenario. However, researchers who want to demonstrate new results on a physical platform face difficult challenges, if they do not have access to a robotic platform in their own labs. Thus, there is a need for a research testbed where simulation-based results can be rapidly validated through hardware in the loop simulation, in order to test the software on board the physical platform. The CAT Vehicle Testbed offers such a testbed that can mimic dynamics of a real vehicle in simulation and then seamlessly transition to reproduction of  use cases with hardware. The simulator utilizes the Robot Operating System (ROS) with a physics-based vehicle model, including simulated sensors and actuators with configurable parameters. The testbed allows multi-vehicle simulation to support vehicle to vehicle interaction. Our testbed also facilitates logging and capturing of the data in the real time that can be played back to examine particular scenarios or use cases, and for regression testing. As part of the demonstration of feasibility, we present a brief description of the CAT Vehicle Challenge, in which student researchers from all over the globe were able to reproduce their simulation results with fewer than 2 days of interfacing with the physical platform. 
\end{abstract}

\section{Introduction}
The last decade has seen a rapid increase in the research interest in autonomous driving technology. An increasing trend towards bringing various levels of autonomy can be seen in the vehicular technology, which has taken an incremental approach (rather than disruptive approach of the DARPA Grand Challenge's demonstrations \cite{behringer2004darpa, chen2004ohio, seetharaman2006unmanned, thrun2006stanley}) to bring autonomy to passenger vehicles. Vehicular automation has slowly moved from self-parking and lane assistance to traffic aware cruise control and conditional automation. In recent years, a number of major industry players have emerged to advance the autonomous driving technology \cite{colquitt2017driverless, rimmer2017intellectual}.
Motor vehicle companies such as BMW, Audi, Bosch, and General Motors have predicted the readiness of vehicles with full automation and its mass production by 2020 and companies like Google and Uber are investing heavily to bring full automation in driving \cite{chan2017advancements}. Research groups in academia and industry have built their own platforms to investigate research problems in autonomous driving \cite{how2008real, hebert2012intelligent, reimer2014assessing}. Despite the advances in autonomous driving technology, high fidelity automation in driving requires testing of autonomous features in every scenario. The design, implementation, and testing of vehicles under a wide range of use cases and in realistic traffic conditions is costly, time consuming, complicated, and often un-reproducible. Integration testing with the physical platform in these cases are unnecessarily complex and often performed in the last stage of development process. This makes prototype design, implementation, intensive testing and simulation with an actual vehicle in the simulation loop the most effective way to verify and validate the design idea.
Software in the loop (SIL) simulation in the laboratory environment offers a safe way to perform prototyping and implementation of vehicle control and algorithms. Incorporating an actual vehicle and sensors in the simulation loop (called hardware in the loop or HIL) validates the design and reduce the time required for the system verification.

\subsection{Contributions}
The contribution of this work is a testbed with HIL simulation that offers an integrated architecture for investigating, prototyping, implementing, and testing a concept in autonomous driving technology. Our testbed allows researchers to examine the performance of sensor-guided driving in a repeatable fashion for safe, flexible and reliable validation. The main contributions can be summarized as follows:
\begin{itemize}
\item An open-source, experimentally validated and scalable testbed with HIL support for autonomous driving applications that uses ROS.
\item A simulated physics-based model of the autonomous vehicle that mimics a real world vehicle capable of driving autonomously.
\item A multi-vehicle simulator that provides a virtual environment capable of testing a research application requiring vehicle to vehicle interaction.
\item A research paradigm that enables distributed teams to implement and validate a proof of concept before accessing the physical platform.
\item Software packages, descriptions of sensors and methodology employed to develop autonomous vehicle applications.
\end{itemize}

\subsection{Related Works}
Although autonomous driving has been an area of research interest for a long time, the DARPA Grand Challenge inspired research community to develop a number of autonomous vehicle testbeds across the academia and the industry. Stanford's Junior \cite{levinson2011towards} provides a testbed with multiple sensors for recognition and planning. It is capable of dynamic object detection and tracking and precision localization. Other few notable testbed born as a result of DARPA challenge are Talos from MIT \cite{leonard2008perception},  NavLab11 and Boss from CMU \cite{urmson2008autonomous}. Costley et al. discuss a testbed for automated vehicle research available in Utah State University \cite{costley2017low}. Researchers at University of California, San Diego have also developed a testbed named LISA-Q \cite{mccall2004design} for intelligent and safe driving. Availability of testbeds for vehicle research is not limited to ones mentioned here, but to the authors' knowledge they lack extensive support for HIL simulation. Moreover, none of the previous works mention any physics-based simulator offering support for multi-vehicle simulation.

In this paper, we describe the CAT Vehicle (Cognitive and Autonomous Test Vehicle) Testbed, which is comprised of the Robot Operating System (ROS) \cite{quigley2009ros} based simulator that takes advantage of the ODE physics engine \cite{ODE}. The testbed provides packages for a number of control applications and sensor simulations, and interfaces through which new custom packages can be installed. The testbed is designed to support code generation, in which researchers can transfer their MATLAB or Simulink \cite{codegensimulink} implementations from simulation to the physical platform through a straightforward workflow. The testbed is further equipped with a tool to visualize data being streamed by either real hardware such as an actual autonomous vehicle (AV) and sensors or simulated AV and sensors, through straightforward reconfiguration.

Further, in this paper, we describe the architecture of the testbed, software tools, vehicle dynamics and supported sensors. We further describe the physical platform for the implementation, communication channels, visualization for diagnostics and debugging, and data. At the end, we discuss how the testbed enables a research paradigm in which distributed teams validate and test their work before they have access to the physical platform. In the subsequent stage, teams are then able to demonstrate their results on the physical platform in the span of 2 days or less. We conclude with lessons learned with our development and outlook for the future work.

\section{Overview of Testbed Architecture}
\begin{figure}[htbp]
\centering
\includegraphics[width=0.9\textwidth]{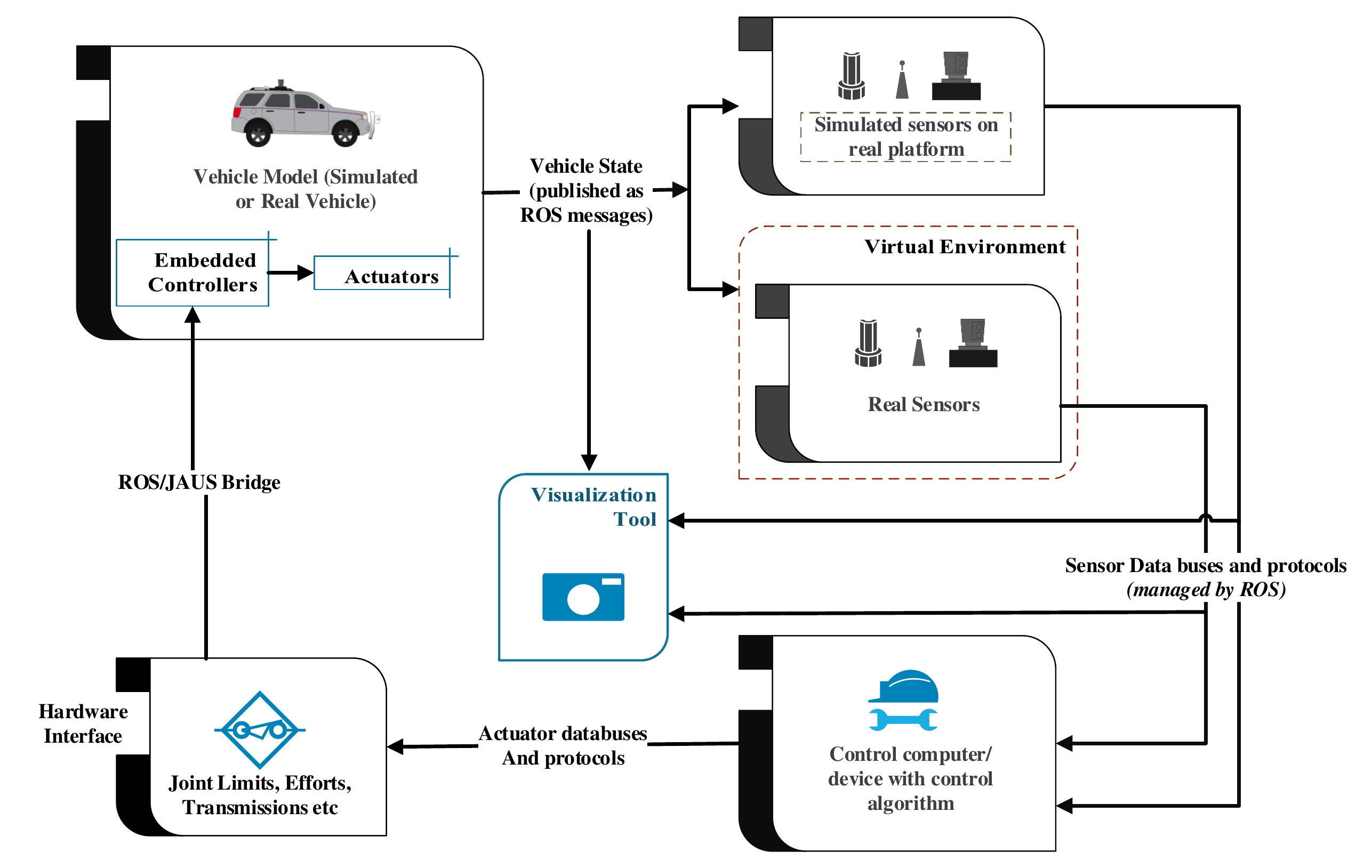}
\caption{A block diagram showing important components in the CAT Vehicle Testbed. Dotted line around real sensors denotes that a virtual environment can host real sensors.}
\label{fig:tesbed_architecture}
\end{figure}

A schematic diagram of the architecture of the CAT Vehicle Testbed is shown in figure \ref{fig:tesbed_architecture}. It supports modular design with an emphasis on scalability. The testbed consists of a virtual environment, a vehicle model, a physics engine, sensor models, a visualization tool, command line tools and scripting tools for interfacing and code generation capability. Each component is discussed in the subsequent sections.

The testbed is based on packages compatible with ROS, using C++ and Python \cite{quigley2009ros}. ROS provides libraries and tools for writing control and perception algorithms, and other applications for autonomous vehicles. With various levels of software and hardware abstraction, device drivers for seamless interface of sensors, libraries for simulating sensors and visualizers for diagnostics purposes, ROS provides a middleware and repository by which distributed simulation can take place, and the software installation is straightforward. Being a distributed computing environment, it implicitly handles all the communication protocol. 

However, standard ROS packages lack domain-specific requirements for experimentation with a car-like robot. A typical setup of autonomous vehicle consists of a vehicle controller and sensors strategically mounted on different points of the vehicle. In order to control motion in a seamless way, we require uniform interfaces for control and consistent consumption of sensor data. The goal of the CAT Vehicle Testbed is to demonstrate how to estimate current state and issue control signals well before the physical platform is engaged. Once algorithms have been demonstrated in SIL, HIL simulation or execution takes place, in which one or all of the physical platform (vehicle and/or sensors) can be replaced by its equivalent simulated version.

Our testbed offers a modular platform with software interfaces to develop and deploy algorithms using model-based design. In this modular platform, there is no distinction between a simulated autonomous vehicle (AV) and a real AV. As a result, the testbed makes deployment and debugging simpler. There are two ways the testbed provides the simplicity for validating and verifying the dynamics of the AV under wide variety of conditions. First, the virtual environment generates synthetic data from the simulated sensors, as if they are from the onboard sensors. These data are relayed to the real AV, which uses the controller designed to produce control commands. The control commands are sent back to the simulated sensors, which close the loop. Alternatively, the real onboard sensors can be used to receive the data and relay to the simulated AV. A controller on the simulated AV receives these sensor data and produces control commands to alter the state of the AV.

\section{Virtual Environment}
The virtual environment consists of an approximate vehicle model, a simulated world and simulated sensors. The most important component that creates a virtual environment is the simulator. In the CAT Vehicle Testbed, the virtual environment uses the Gazebo simulator, which utilizes the ODE physics engine to simulate laws of physics. The Gazebo models are specified to approximate the true physical behavior by tuning friction coefficients, damping coefficients, gravity, buoyancy, etc. 

Our virtual environment enables multi-vehicle simulation, which can be used to simulate vehicle to vehicle interaction, create a leader-follower scenario and traffic like situations. The vehicle model used in the CAT Vehicle Testbed is a kinematics model, which uses simulated controllers to actuate joints of the autonomous vehicle. It uses a ROS-based control mechanism through Gazebo to send control inputs to the car.

\subsection{Vehicle Model}
The complexity of developing a realistic car-like robot in simulation is significant due to a few contributing factors: (1) runtime solvers approximate motion based on constraint satisfaction problems, which can be computationally expensive if the vehicle model's individual components are unlikely to approximate physical performance; (2) kinematic robotic simulation typically utilizes joint-based control, rather than velocity based (or based on transmission/accelerator angles and settings) like a physical platform; and (3) the dynamics of individual vehicle parts is such that physically unrealistic behavior may emerge, meaning that physical approximations of linear and angular acceleration should be imposed on individual joints, to prevent unlikely behaviors. The following section discusses these details.

\begin{figure}[htbp]
\begin{subfigure}{0.4\textwidth}
\centering
\includegraphics[width=0.75\textwidth]{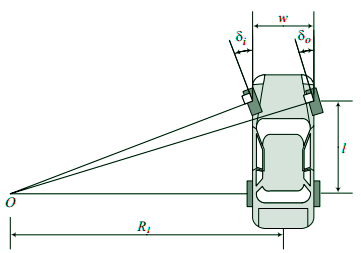}
\captionsetup{justification=centering, font=scriptsize}
\caption{The steering mechanism used in the vehicle model uses Ackermann steering principle: $cot(\delta_0) - cot(\delta_i) = w/l$. The vehicle receives single value of steering angle as input which is calculated as $cot(\delta) = \frac{cot(\delta_0) + cot(\delta_i)}{2}$.}
\label{fig:ackermann}
\end{subfigure}
\begin{subfigure}{0.6\textwidth}
\centering
\includegraphics[width=0.6\textwidth]{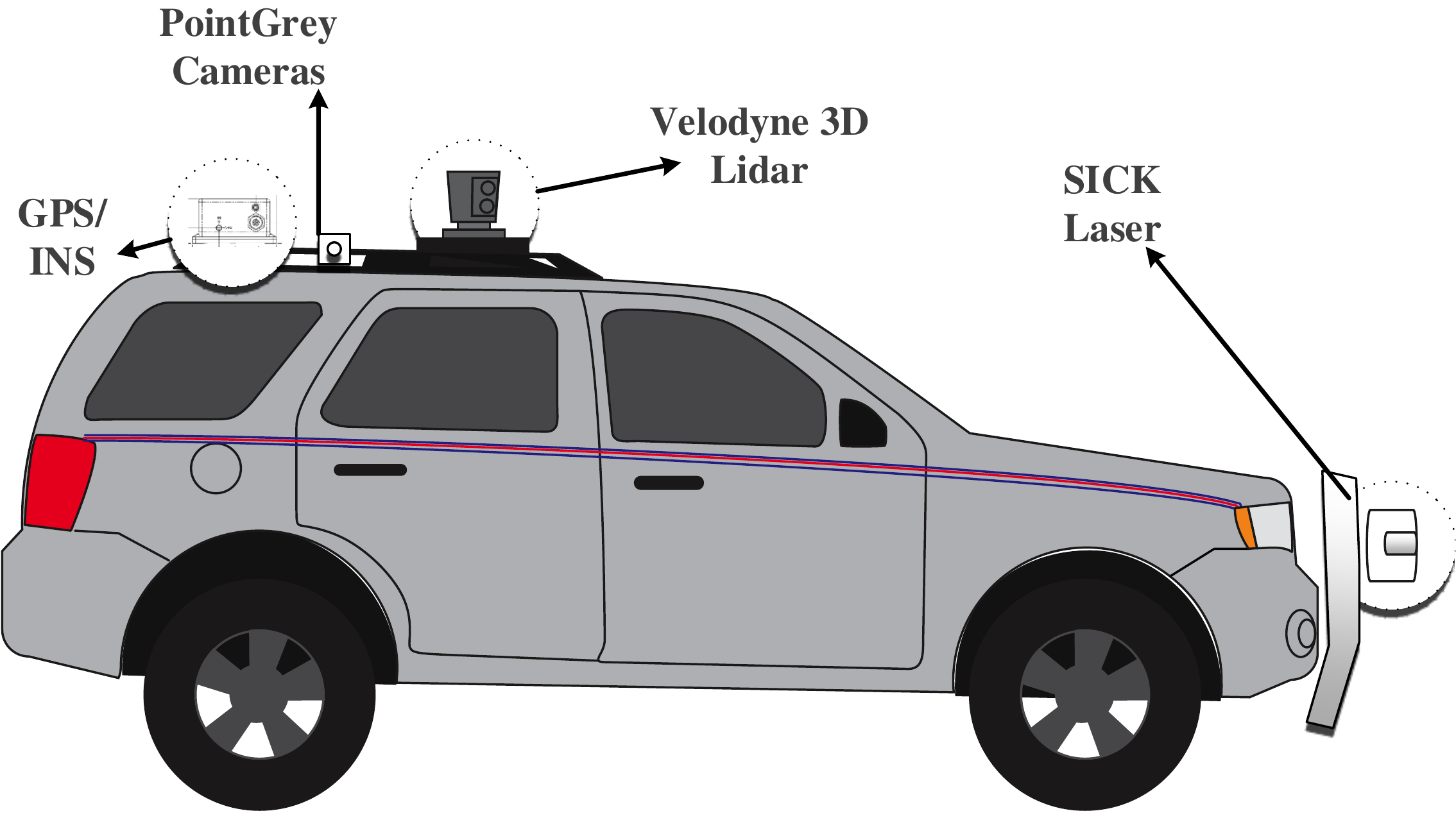}
\captionsetup{justification=centering, font=scriptsize}
\caption{A schematic diagram of vehicle model \\showing positions of different sensor mounted on it.}
\label{fig:vehicle_model}
\end{subfigure}

\captionsetup{justification=centering}
\caption{}
\end{figure}

In the vehicle model (figure \ref{fig:vehicle_model}), left and right rear wheels move the robot using a joint velocity controller, simulating rear-wheel drive for the vehicle. The wheels use a PID-controlled mechanism to meet linear velocity set points, with constraints that prevent unrealistic acceleration (or deceleration). As alluded to above, the physical dynamics of motion on a plane mean that there is typically slippage of tires. In lieu of a limited slip differential (the approach on modern physical cars) to reduce rear tire slip when turning, the velocity controller of the CAT Vehicle utilizes the circle of curvature to geometrically determine a modifier for the desired rotational velocity of each rear tire, depending on the steering angle.

When steering the vehicle, the steering position controller utilizes the Ackermann steering model for the front two tires \cite{thompson2007ackerman, milliken1995race}. The Ackermann steering model states that while turning on a curvature, inner and outer tires need to be turned at different angle to avoid tire slip as shown in figure \ref{fig:ackermann}. If a simulated car-like robot does not account for this physical constraint, then the physics-based solver will return infeasible solutions, and/or result in tremendous slow-down of the computation speed for the solver. Steering values are set based on expected dynamics of a realistic steering controller; this results in a time delay between set angles, to reflect the time taken to steer from one angle to another. These dynamics are set in the steering controller, and can be configured.

The structure of the vehicle model is represented using the Unified Robot Description Format (URDF), which is in the form of XML Macros (xacros). An URDF file contains description of the visual geometry and configuration of different parts of the vehicle and specifies how different parts are connected together. Furthermore, this file specifies where simulated sensors will be mounted on the vehicle. This file also contains information about inertial properties of different parts of the vehicle and collision geometry (different from its visual geometry). We also specify actuator and transmission configuration in the URDF file. Physical attributes such as kinematics and dynamics constraints on the different components of the vehicles such as coefficients of frictions and damping coefficients are specified in a SDF format file end with extension \texttt{.gazebo}. This file also contains additional plugins that bind to the vehicle model to provide specific functionality.

\subsection{Simulated World}
\begin{wrapfigure}[13]{r}{0.45\textwidth}
\centering
\includegraphics[width=0.45\textwidth]{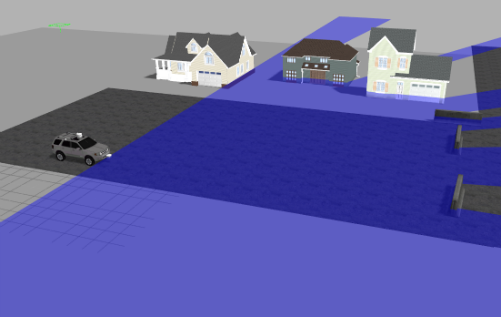}
\caption{A simulated world, with front-mounted laser data visualized in Gazebo.}
\label{fig:gazebo_world}
\end{wrapfigure}

The virtual world or simulated world provides a virtual reality where a simulated vehicle and its sensors interact with each other and with virtual objects rendered in the virtual world. The CAT Vehicle Testbed uses Gazebo and built-in ODE physics engine to render the virtual reality. The configuration of the virtual world can modify the simulation step size, which is the maximum time step by which every system in the simulation interacts with states of the world. One can also specify real time factor to slow down or speed up the simulation. The \texttt{.world} also offers a way to adjust real time update rate at which physics engine update the state of the simulation. Some aesthetical aspects such as look and feel, geometric orientation of the world and lighting can also be adjusted. An example of simulated world in Gazebo is shown in figure \ref{fig:gazebo_world}. Importantly, the world file permits users to develop approximations of the desired HIL verification environment, so that the SIL testing can be shown to work in an approximation of the physical environment. Various components from Gazebo can be used to add pavement, signs, traffic cones, etc., in order to more closely approximate the expected HIL scenarios.

\subsection{Simulated sensors}
The CAT Vehicle Testbed offers three types of simulated sensors: Front Laser Rangefinder, Velodyne LIDAR, and two side cameras. The Front Laser Rangefinder simulates the behavior of commercially available SICK LMS 291 Rangefinder, which is also a part of physical platform available with the CAT Vehicle Testbed. The Front Laser Rangefinder can detect an object at a distance of upto 80 m and scans its surrounding from 0 to 180 degree at a rate of 75 Hz. The angular resolution of the Rangefinder is 0.0175 radian.

The Velodyne LIDAR produces point cloud data that can be analyzed to obtain three-dimensional information about its surroundings. Our simulated LIDAR only provides one measurement channel. Hence, it does not provide any color information. Point cloud data from the Velodyne LIDAR is published at a rate of 5 Hz by default. In the CAT Vehicle Testbed, the Velodyne LIDAR is configured for 100 horizontal beams and 20 vertical beams. By default, each horizontal beam has minimum angle of -0.4 radian and maximum angle of 0.4 radian. Each vertical beam has angular range of -0.034906585 to 0.326 radian.  With a resolution of 0.02m, the Velodyne LIDAR can detect objects as far as 50 meters from its position. Although, these values are default, they are potentially reconfigurable depending on the use case. As a result of its fine resolution, the volume of data produced by the LIDAR is in the order of Gigabytes per minute. Hence, caution must be taken while using the Velodyne LIDAR data, otherwise data channels can be flooded with point cloud data and other packets might be dropped altogether.

Two cameras are mounted on the left and right side of the vehicle. Each Camera produces images of the size 800x800 at a rate of 30 Hz in the RGB 8 bit format. They have horizontal field of view of 1.3962634 radians. ROS offers different data types to obtain camera images in raw, RGB and grayscale format. Outputs from these cameras can be consumed by standard computer vision toolboxes such as Simulink toolboxes or OpenCV.

\section{Physical Platform}
\begin{figure}[htbp]
\begin{subfigure}{0.5\textwidth}
\centering
\includegraphics[width=0.65\textwidth]{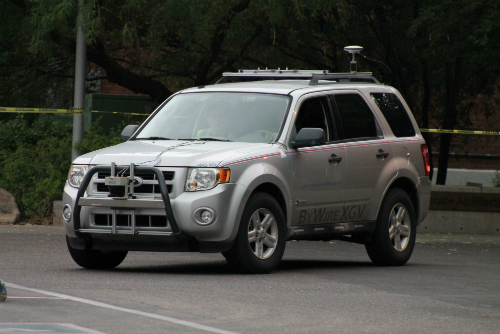}
\captionsetup{justification=centering, font=scriptsize}
\caption{The CAT Vehicle with GPS/IMU\\and Laser Rangefinder mounted on it.}
\label{fig:catvehicle_physical}
\end{subfigure}
\begin{subfigure}{0.5\textwidth}
\centering
\includegraphics[width=0.68\textwidth]{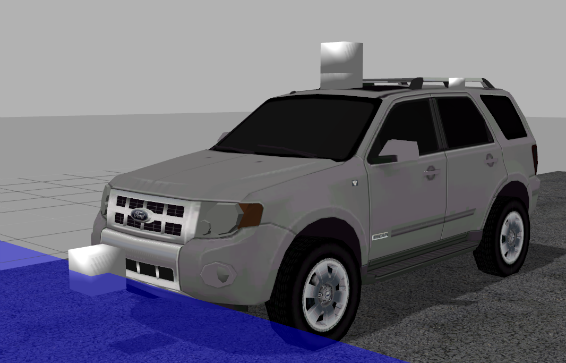}
\captionsetup{justification=centering, font=scriptsize}
\caption{The model of the CAT Vehicle\\used in the simulation}
\label{fig:catvehicle_gazebo}
\end{subfigure}

\captionsetup{justification=centering}
\caption{Left: The the autonomous vehicle used in the CAT Vehicle Testbed.\\Right: An equivalent 3D model in the simulation.}
\end{figure}

The physical platform used for the CAT Vehicle HIL is a modified Ford Hybrid Escape vehicle (Figure \ref{fig:catvehicle_physical}), upon which is mounted a SICK LMS 291 Front Laser Rangefinder, a Velodyne HDL-64E S2 LIDAR, two Pointgrey Firefly MV FFMV-03M2C cameras and a Novatel GPS/IMU. The sensors are collectively called as the perception unit.

\subsection{The Autonomous Vehicle in the Simulation Loop}
The physical testing platform used for autonomous vehicle application is TORC ByWire XGV drive-by-wire platform with modified Ford Hybrid Escape vehicle. This physical platform consists of multiple hardware subsystem control modules, a central embedded controller and a TORC SafeStop ES-220 multilevel wireless emergency stop system designed to command pause and stop states in the event of an emergency. This vehicle utilizes the Joint Architecture for Unmanned Systems (JAUS) \cite{wade2006joint} as the standard interface to receive state data from the vehicle and send controller inputs to the vehicle. All communication with the computer that manages the low-level actuation and control of the vehicle requires commands delivered through JAUS. Since our virtual platform uses ROS, we created a ROS/JAUS bridge to create a layer of abstraction to facilitate sending control command to vehicle using the ROS \cite{morley2013generating}. Use of this bridge made integration with the ROS-based Gazebo simulator feasible. The onboard computer of the CAT Vehicle runs Ubuntu 14.04, which has ROS packages and the ROS/JAUS bridge installed for running device drivers to interface with sensors and for control commands. 

\subsection{Perception Unit}
A SICK LMS 291 Laser Rangefinder, a Velodyne HDL-64e LIDAR, two Pointgrey cameras and a Novatel GPS/IMU constitute perception unit. The Laser Rangefinder is mounted on the front bumper of the car at a height of approximately 75 cm from the ground. The Rangefinder gives the distance information in the form of 180 points every 1/75 seconds. It offers the simplest solution to make decisions about steering and collision avoidance. The measurement resolution provided by SICK LMS 291 LIDAR is 10mm. The Velodyne LIDAR is mounted on the top of the vehicle that provides the ability to construct three-dimensional imagery of the surrounding area. It uses 905nm laser beam to provide 360 degrees field of view. Pointgrey cameras can be mounted on the left and the right side of the vehicle to get information about parallel traffic. In the simulation, getting a real time position of the vehicle is trivial which is accomplished physically by using a Novatel GPS/IMU unit. The Novatel GPS/IMU unit is based on Global Navigation Satellite System (GLONASS) and The Global Positioning System (GPS). It has performance accuracy of 1.5m for single point L1/L2, 0.6m for SBAS and 0.4m for DGPS. For the purpose of velocity estimation, accuracy of the GPS/IMU unit is sufficient but for higher accuracy, the system must have a base station that our current infrastructure does not support.

\section{Modeling and Implementation}
\begin{figure}[htbp]
\centering
\includegraphics[width=1.0\textwidth]{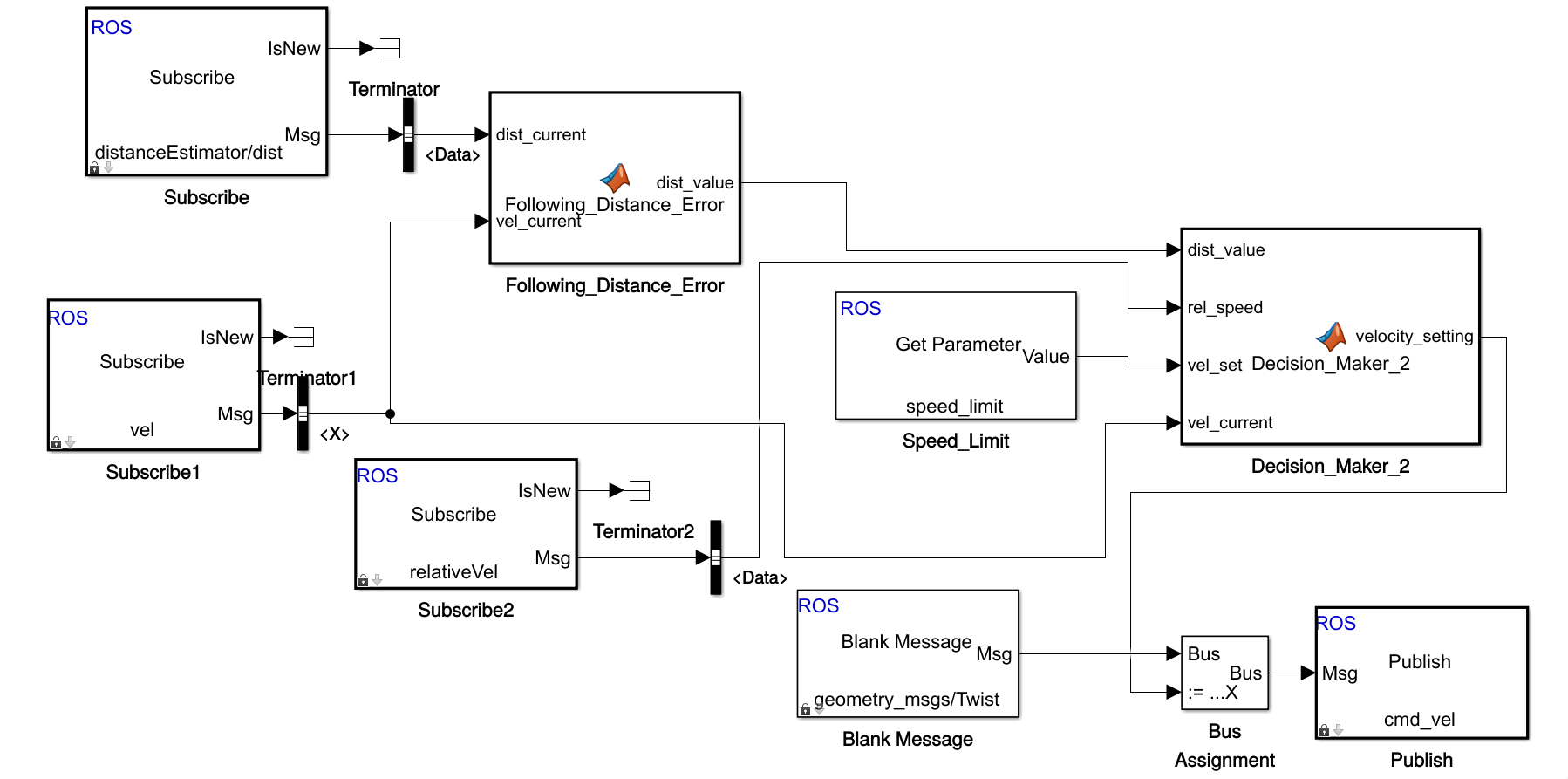}
\caption{A Simulink model for implementation of fuzzy-based car-following algorithm. It has two functional blocks: \textit{Following\_Distance\_Error} and \textit{Decision\_maker\_2}. The first block takes headway distance and instantaneous velocity as an input and calculate a distance error in seconds and feeds to the second block. The second blocks makes the decision about the next step based on the inputs it receives.}
\label{fig:simulink_model}
\end{figure}

ROS-based packages and APIs enable developers and researchers to write applications for autonomous vehicle in modular fashion to interact with simulation and physical platform. The CAT Vehicle Testbed uses C++ and Python APIs available from the ROS open source community for prototyping,  designing and testing control algorithms and solutions for autonomous vehicles. Each component (node) of ROS can receive and send data from and to other nodes by virtue of the distributed communication feature of the ROS. Each ROS node can be reused and instantiated multiple times, thereby facilitating multi-vehicle simulation. Command line and scripting tools from ROS enables developer to define relationships between different components and develop custom solutions. In some cases, handling and processing of complex data such as point cloud data from a LIDAR using C++ and Python packages prove to be unnecessarily complex and substantially slow down the development process. The Robotics System Toolbox in MATLAB/Simulink provides a component-based solution where existing components from other domains such Control Systems, and Computer Vision can be reused with the Robotics Systems Toolbox to speed up the development time \cite{lizarraga2009simulink, gans2009hardware}. Simulink provides model-based design with the drag and drop feature. One such Simulink model is shown in figure \ref{fig:simulink_model}. Its code generation facility allows generation of ROS packages in the form of C++ artifacts that can be integrated with the existing testbed.
Users can export generated C++ code to an existing ROS workspace, for use with either the simulator or HIL. They can be run on either the CAT Vehicle’s onboard computer or bound with the simulated AV.

During a simulation, the Gazebo simulator publishes the sensor data from the simulated world that can be subscribed by the velocity controller, either implemented in the physical vehicle or attached with the simulated vehicle. Based on the sensor data, the velocity controller produces the required command that is consumed by a ROS-based controller and translated to inputs for the actuators. In this process, a ROS-based controller does not distinguish between a simulated vehicle and an actual vehicle. Actuators in the actual vehicle compute the forces and torques required to achieve the desired velocity and accordingly apply brake pressure or throttle.

The testbed also offers a multi-vehicle simulation to study vehicle to vehicle interactions, leader-follower scenarios, and to explore novel traffic model. Our testbed provides ready-to-use configuration files that can be used to spawn new vehicle models in the simulator on the go. These newly spawned vehicle models can be given different velocity profiles to simulate human driving behaviors or act as another autonomous vehicle. A user can drive these simulated vehicles (or even the real CAT Vehicle) using a PS2 joystick.

\subsection{Data}
In the initial stage of testing a prototype, a physical vehicle is almost never used for safety reasons. ROS provides command line tools to capture date-stamped data published by other nodes in the system (such as the onboard computer from the CAT Vehicle) in the form of \texttt{bag} files. These \texttt{bag} files may contain data from actual sensors such as GPS/IMU, LIDAR or Rangefinder, as well as vehicle data such as its velocity, brake pressure and throttle. These bag files may be played back in real time in the simulation loop to reconstruct the exact scenario against which a controller prototype can be tested and validated. Bag files collected from the sensors while driving the CAT Vehicle can also be used for regression testing, verification and debugging.

\subsection{Visualization}
Logging sensor information and visualization of data is an imperative part of the development and debugging processes. 
ROS provides \textit{Rviz} as a visualization tool for debugging and diagnostics purposes. The CAT Vehicle Testbed provides a custom configuration that can be loaded in 
\begin{wrapfigure}[15]{r}{0.48\textwidth}
\centering
\includegraphics[width=0.48\textwidth]{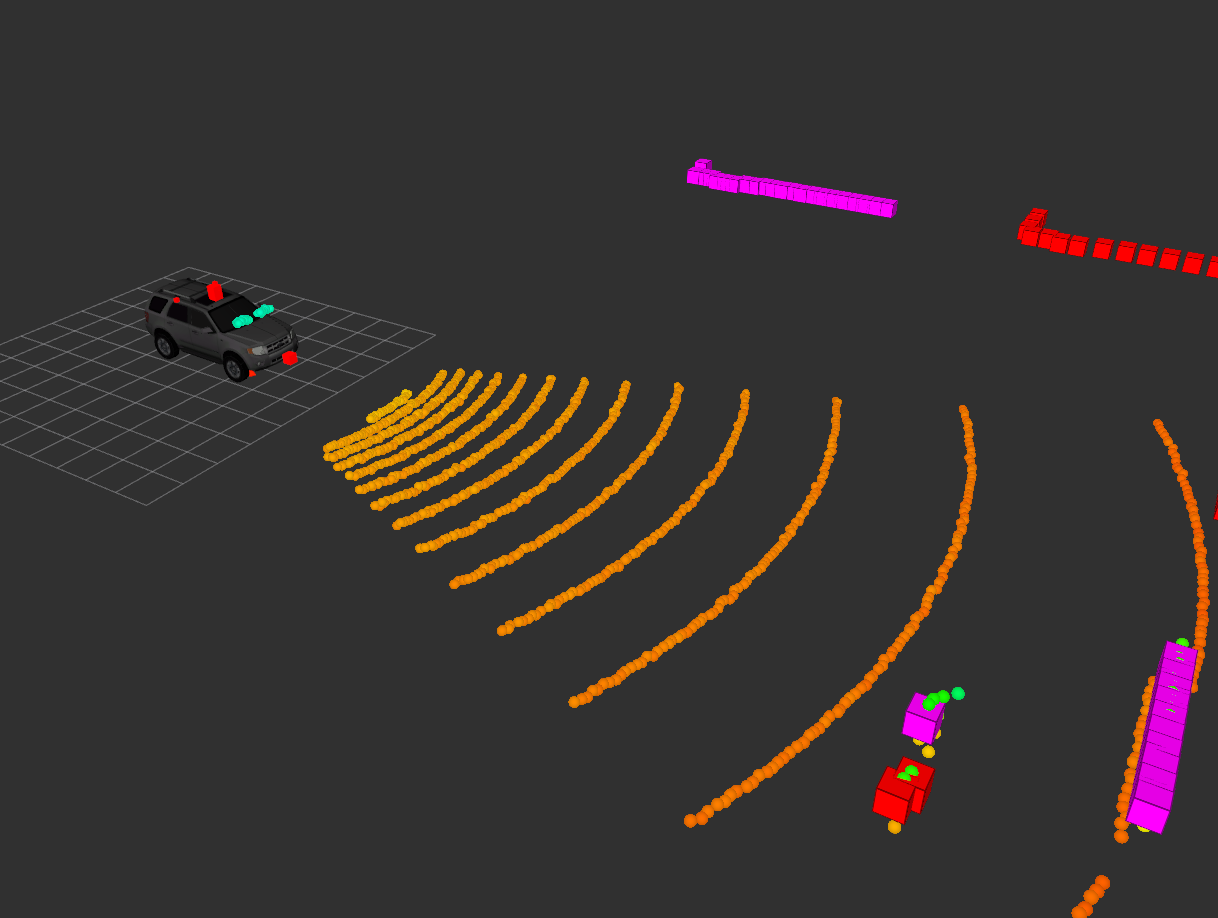}
\vspace{-10pt}
\caption{Rviz visualization.}
\label{fig:rviz_rviz}
\end{wrapfigure}
 \textit{Rviz} to visualize a path traced by either a simulated vehicle or an actual vehicle, LIDAR pointcloud data, rangerfinder data and camera data. \textit{Rviz} provides an ability to change the frame of reference from one component to another thereby providing a mechanism to test correct kinematics and dynamics of the different components involved in the autonomous vehicle applications. Figure \ref{fig:rviz_rviz} demonstrates one such visualization obtained during our experiments. Visualization can be performed in real time by suitable coordinate transform, or by playing back recorded data from bag files in real-time at either slower- or faster-than-real-time.

\begin{figure}[htbp]
\begin{subfigure}{0.5\textwidth}
\centering
\includegraphics[width=0.9\textwidth]{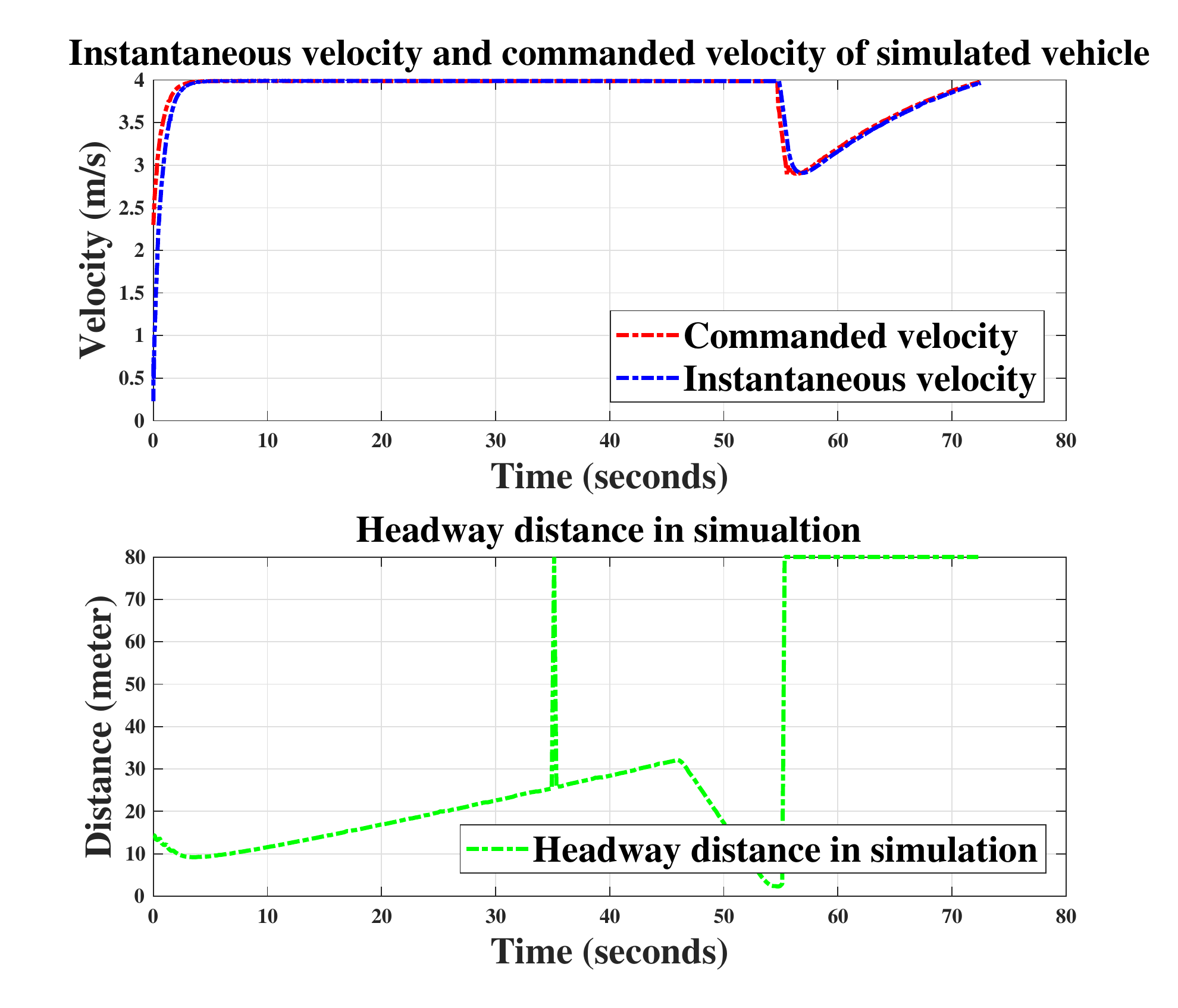}
\captionsetup{justification=centering, font=scriptsize}
\caption{Velocity and headway \\ profile from the pure simulation}
\label{fig:ml_controller_simulation}
\end{subfigure}
\begin{subfigure}{0.5\textwidth}
\centering
\includegraphics[width=0.9\textwidth]{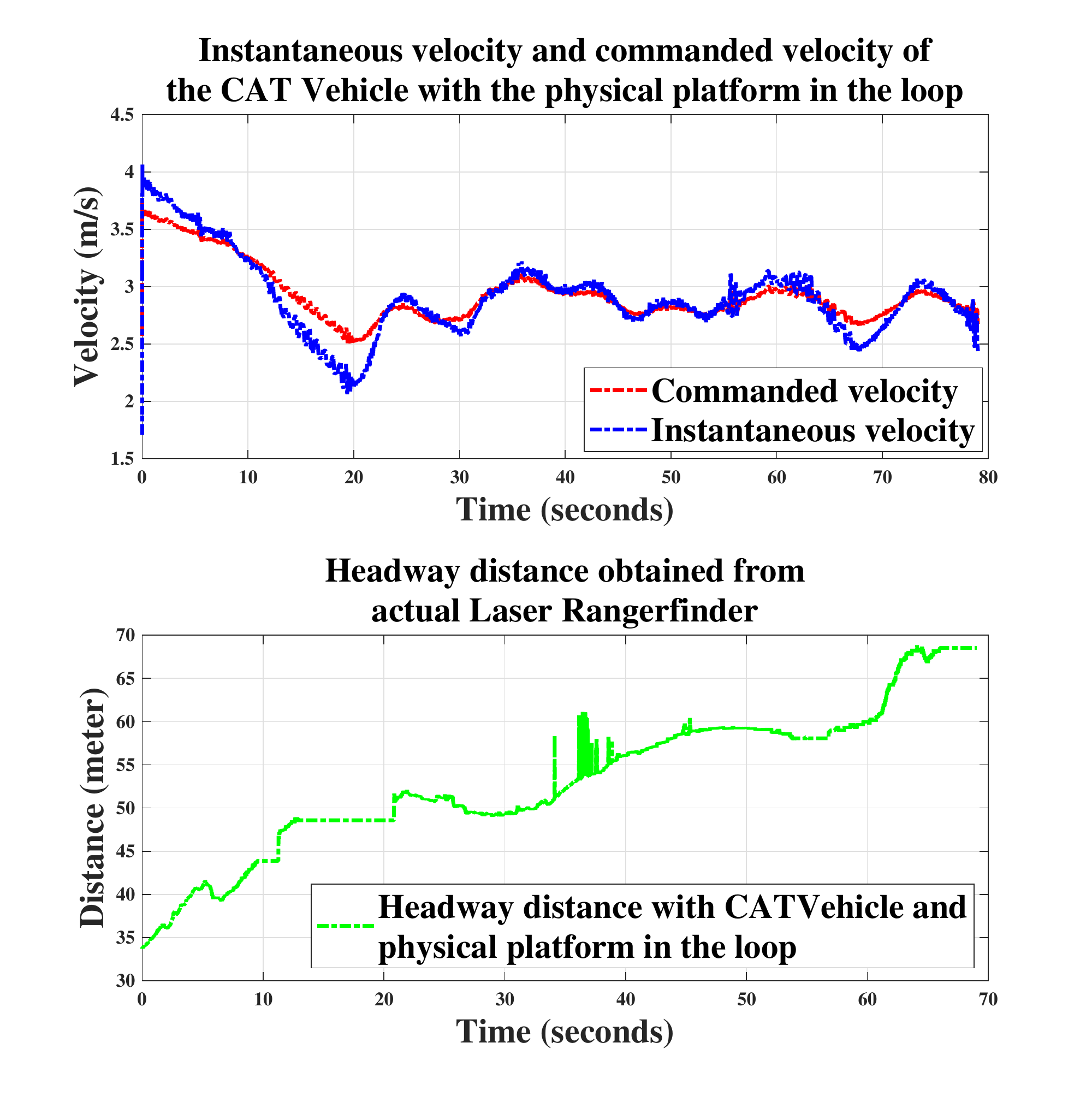}
\captionsetup{justification=centering, font=scriptsize}
\caption{Velocity and headway \\ profile from physical platform}
\label{fig:ml_controller_HILS}
\end{subfigure}

\captionsetup{justification=centering}
\caption{Both plots were generated from data obtained as a result of running the same controller, first in a simulation and then on the physical platform. In the lower subplots, headway distance are presented in its raw format without any smoothing. Spikes in headway distance are results of curvature in the vehicle trajectory and deviation of road surface from perfect flatness.}
\end{figure}

Figure \ref{fig:ml_controller_simulation} and \ref{fig:ml_controller_HILS} presents velocity and headway profile obtained from the testbed with the same controller ran first in the simulation and then with the physical platform without any need for further modification. The controller used in this example was a velocity controller that implements a vehicle-follower algorithm. In the simulation, the preceding vehicle was given a constant velocity profile as a command input whereas in the physical platform demonstration a human was driving the preceding vehicle. The controller used for this simulation was designed to send command inputs to the autonomous vehicle to follow its preceding vehicle.

\subsection{System Safety}
\label{sec:safety}
From the inception of an idea to its final prototype and testing, safety of vehicle applications have remained our prime concern. HIL simulation mitigates the risk of failure or unintended action of controllers under test by extensive testing in the virtual environment with synthetic as well as real data and a combination of simulated and real sensors. The testbed architecture ensures the validity of controller requirements and design with multiple testing with SIL simulation followed by HIL simulation. Thus, the CAT Vehicle Testbed provides safety by allowing researchers to design safe system dynamics.

Considering that autonomous vehicle applications consist of high-risks scenario, we provide a layer of package for collision avoidance called as \textit{obstaclestopper}. \textit{Obstaclestopper} guarantees safety by continuously monitoring the minimum distance calculated from the Rangefinder data. As needed, it modifies the desired velocity commanded to the AV to prevent any potential collision. Further, as discussed in section the \ref{sec:DSML}, we have shown that another layer of abstraction can be created to ensure functional correctness of the system with the help of the CAT Vehicle Testbed. In addition to that, the physical platform consists of an emergency stop that can be utilized in an unforeseen event to prevent any damage to life and property.

Although the testbed offers various degrees of permissive safety and functional correctness, it does not offer any kind of security since the testbed uses ROS for message communication. In ROS network, anyone with the information about IP address of ROS master can access messages being exchanged by subscribing to specific topics. Any other process can kill a node in the network and a new process with the same name can supersede another node \cite{secureROS2017}. Further, node-to-node communication are done in plain text which is vulnerable to spoofing \cite{mcclean2013preliminary}. As a result, any sensor data are prone to manipulation and even failure which can result in catastrophic situations.
\section{Enabled Research and Applications}
During its development process, the CAT Vehicle Testbed has enabled research progress in Domain Specific Modeling Languages (DSMLs), traffic engineering and intelligent transportation systems, visualization and perception applications of sensor data, and promoting STEM education and robotics among school kids. In the following subsections, we present a brief summary of research works and projects that employed the CAT Vehicle Testbed.

\subsection{Dissipation of stop-and-go traffic waves}
\begin{figure}[htbp]
\centering
\includegraphics[width=0.60\textwidth]{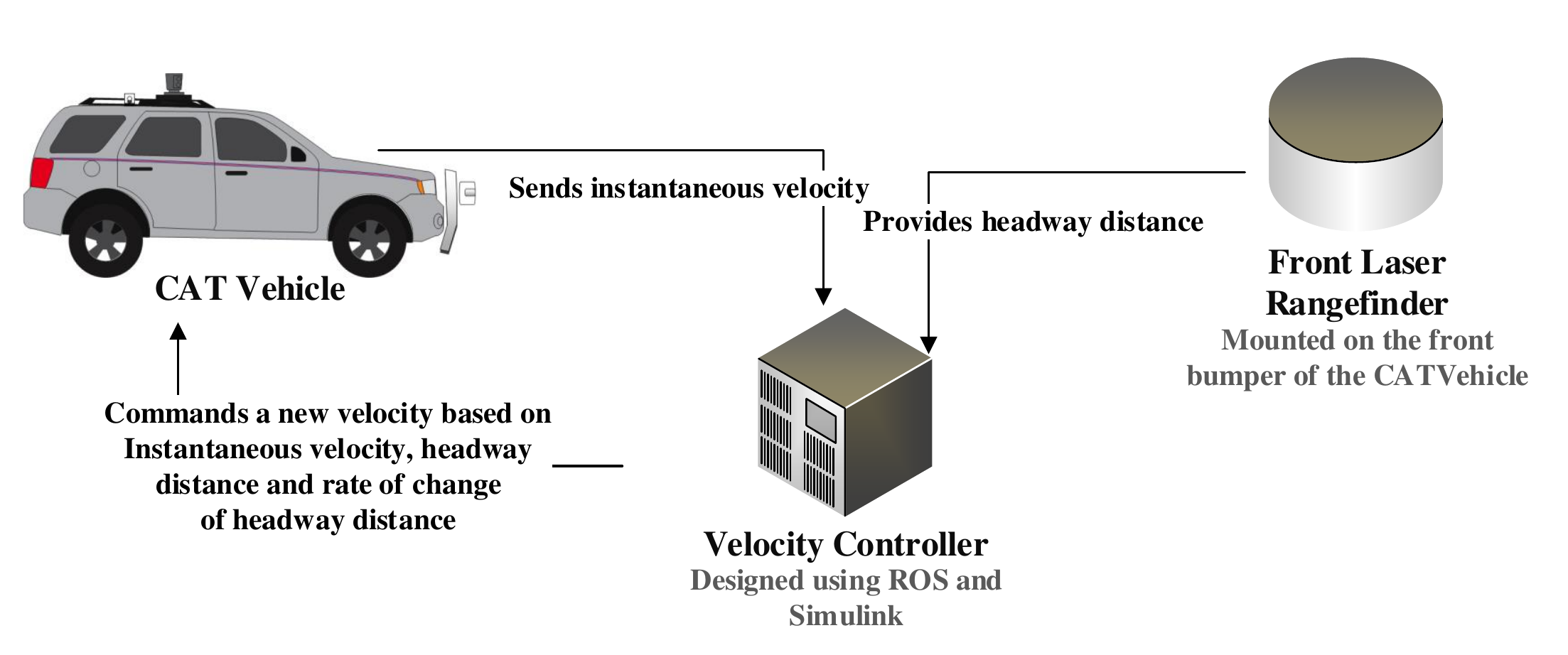}
\caption{The top-level schematic of velocity controller used to dampen emergent traffic waves.}
\label{fig:velocity_controller}
\end{figure}

In 2016, we conducted an experiment to reproduce Sugiyama experiment \cite {sugiyama2008traffic} with some mdodifications to suit common driving conventions in the United States. In this experiment, we demonstrated that intelligent control of an autonomous vehicle can be used to dissipate stop-and-go traffic waves that can arise even in the absence of a physical condition or a lane change \cite{stern2017dissipation}. In this experiment, the CAT Vehicle Testbed played an imperative role in proving the hypothesis. The velocity controller used in dampening the traffic waves was designed using MATLAB/Simulink and Robotics System Toolbox, and utilized extensive bagfile-based regression testing, and component-based testing, in order to take advantage of the dynamically-realistic simulator. The top-level schematic of the controller can be seen in figure \ref{fig:velocity_controller}.

\subsection{Domain Specific Modeling Language for non-experts}
\label{sec:DSML}
\begin{figure}[htbp]
\begin{subfigure}{0.5\textwidth}
\centering
\includegraphics[width=0.8\textwidth]{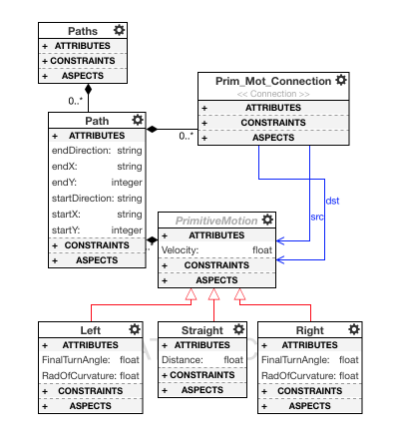}
\captionsetup{justification=centering, font=scriptsize}
\caption{The metamodel for designing paths.}
\label{fig:dsml_meta}
\end{subfigure}
\begin{subfigure}{0.5\textwidth}
\centering
\includegraphics[width=1.0\textwidth]{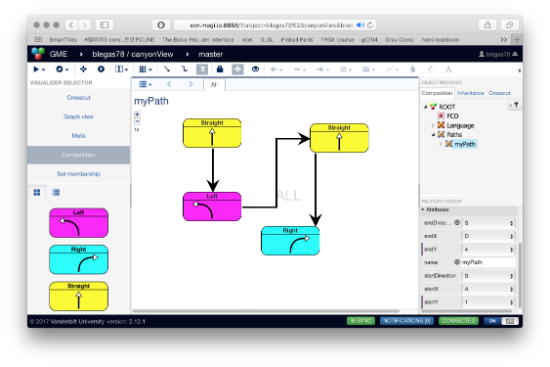}
\captionsetup{justification=centering, font=scriptsize}
\caption{The modeling interface that 4th graders used to operate an autonomous vehicle in non-expert fashion.}
\label{fig:dsml_interface}
\end{subfigure}
\captionsetup{justification=centering}
\caption{}
\end{figure}

An autonomous vehicle is an example of a system that requires experts from multiple domains to collaborate to provide a safe solution. A higher level of abstraction provides a verification tool to domain experts to check for dynamic behavioral constraints. With proper modeling and verification tools, even non-experts can program custom behaviors while ensuring that the system behaves correctly and safely. Using the CAT Vehicle Testbed and WebGME \cite{tunc2016web}, we created a domain specific modeling language (DSML) with a focus on non-experts. The domain in our work consists of driving a vehicle through a set of known waypoints by means of multiple simple motions in a specified manner. Model verification was implemented to ensure safety of the platform, and validation of the problem solution. The language was then provided to a group of 4th grade students to create unique paths. Models created by these children were used to generate controller artifacts and operate the CAT Vehicle on a soccer field \cite{bunting2016safe}. Figure \ref{fig:dsml_interface} shows the modeling interface that was used by kids to design paths for the CAT Vehicle to follow. Figure \ref{fig:dsml_meta} shows the meta model of DSML that the modeling interface uses to generate C++ artifacts for the CAT Vehicle.

\subsection{CAT Vehicle Challenge}
In the spring of 2017, the CAT Vehicle challenge took place to broaden participation in Cyber-Physical Systems. The competition was conducted in four phases from January 2017 to April 2017 \cite{CATVehicleChallenge}. We provided a version of the CAT Vehicle simulation environment, tested by the participants with data obtained from driving the CAT Vehicle. The goal of the CAT Vehicle Challenge was to use data to classify potential obstacles along and off the path, with synthesis of a realistic world file at the conclusion of driving the vehicle. 

During the first three stages of the competition, participants were unable to access the physical platform. They used the CAT Vehicle Testbed with the software in the loop simulation and validated their algorithms and model, submitting bagfile artifacts from their simulation in order to qualify for continued performance in the challenge. In the fourth and final stage, selected participants were invited to the University of Arizona campus to work with the physical platform. 

Each team selected in the final stage of the competition was able to verify their model with the CAT Vehicle and real sensors within the span of two days. They successfully demonstrated their controllers in action by operating the CAT Vehicle autonomously. The result of the final stage was a configuration file for Gazebo simulator to create a virtual world to mirror the environment from which the data was extracted. The CAT Vehicle Challenge proved to be a remarkable example of the research paradigm that employs software and hardware in the loop simulation for design, verification and validation of applications for autonomous vehicles.

\subsection{CAT Vehicle REU Program}
Since its inception in 2014, the CAT Vehicle Testbed has grown alongside CAT Vehicle REU (Research experience for undergraduates) program. Every year, the CAT Vehicle Testbed and CAT Vehicle REU program has contributed to make each other stronger. Each summer CAT Vehicle REU program recruits undergraduates from different institutions across the United States who work under the mentorship of PhD students and professors to advance the autonomous driving technology. Participants in the REU program have made significant contribution to the number of projects in DSML, vehicle perception algorithms, velocity controllers, use of cognitive radio in autonomous vehicle applications and automotive radar \cite{olson2017, McKeeverExperience2015, sprinkle274, sprinkle271,sprinkle261}.

\section{Conclusion and Outlook}
This paper has described a testbed that has enabled progress in autonomous driving technology and STEM education during the last 4 years. The CAT Vehicle Testbed is unique in the sense that it offers seamless transfer of controller design from simulated environment to physical platform without rewriting any component of the controller. It features an autonomous vehicle that is based on open source packages widely used by robotics community.  With the spirit of open source principle, the CAT Vehicle Testbed is available open source on CPS-VO website and GitHub to contribute towards the progress of the autonomous vehicle technology \cite{CPS_VO_CATVehicle_testbed}.

Although the CAT Vehicle Testbed has proven to be remarkably useful in autonomous vehicle applications, there is much work needed to be done to improve its scope and functionality. In its current state, the CAT Vehicle Testbed is limited to testing velocity-based controllers. The testbed does not provide any interface for acceleration, brake or throttle input. This is in par with the physical platform where the CAT Vehicle can only take velocity input as the control command in its autonomous mode. Currently the testbed is based on Ubuntu 14.04, ROS Indigo and Gazebo 2.0.  The future release of the testbed will target newer versions of Ubuntu, ROS and Gazebo. For incorporating security aspects, we may be looking to adopt Secure ROS \cite{secureROS2017} to mitigate the vulnerabilities discussed in \ref{sec:safety}. We will also be looking to create different flavors of interfaces to test wide variety of controllers.

\section{Acknowledgments}
Support for this project was provided by the National Science Foundation and the Air Force Office of Scientific Research under awards 1253334, 1262960, 1419419, 1446435, 1446690, 1446702, 1446715  1521617. The authors express their thanks to the many REU Students who contributed to the codebase that makes up the CAT Vehicle Testbed, especially Sam Taylor, Alex Warren, Kennon McKeever, Ashley Kang, Swati Munjal and a former PhD student Sean Whitsitt. The authors also express their gratitude to Mathworks for sponsoring CAT Vehicle Challenge.

\bibliographystyle{eptcs}
\bibliography{generic}
\end{document}